\documentclass[runningheads]{llncs}
\usepackage{times}
\usepackage{latexsym}

\usepackage[T1]{fontenc}
\usepackage{hyperref}
\usepackage{color}

\usepackage[utf8]{inputenc}

\usepackage{microtype}

\usepackage{inconsolata}

\usepackage{graphicx}

\usepackage{subcaption}
\usepackage{booktabs} 
\usepackage{multirow} 

\usepackage{algorithm}
\usepackage{algpseudocode}

\usepackage{amsmath}  
\usepackage{amsfonts} 
\usepackage{amssymb}
\usepackage{mathtools} 
\usepackage{wrapfig}

\newcommand{\bW}{\mathbf{W}}
\newcommand{\bA}{\mathbf{A}}
\newcommand{\bB}{\mathbf{B}}
\newcommand{\bx}{\mathbf{x}}
\newcommand{\bX}{\mathbf{X}}
\newcommand{\bY}{\mathbf{Y}}
\newcommand{\bU}{\mathbf{U}}
\newcommand{\bV}{\mathbf{V}}
\newcommand{\bSigma}{\mathbf{\Sigma}}
\newcommand{\bcalT}{\mathbf{\mathcal{T}}}

\begin{document}
\title{DoTA: Weight-Decomposed Tensor Adaptation for Large Language Models}

\titlerunning{DoTA: Weight-Decomposed Tensor Adaptation for LLMs}

\author{
    Xiaolin Hu\inst{1}\thanks{Equal contribution. Work done during the internship at XiaoMi.} \and
    Xiang Cheng\inst{1}\thanks{Equal contribution.} \and
    Peiyu Liu\inst{2} \and
    Wei Liu\inst{3} \and
    Jian Luan\inst{3} \and 
    Bin Wang\inst{3} \and
    Yong Liu\inst{1}\thanks{Corresponding author.}
}

\authorrunning{X. Hu et al.}

\institute{
    Gaoling School of Artificial Intelligence, Renmin University of China 
    \and
    University of International Business and Economics, China \and
    XiaoMi AI Lab, China \\
    \email{\{xiaolinhu, chengxiang1, liupeiyustu, liuyonggsai\}@ruc.edu.cn} \\
    \email{\{liuwei40, luanjian, wangbin11\}@xiaomi.com}
}

\maketitle       

\begin{abstract}
    Low-rank adaptation (LoRA) reduces the computational and memory demands of fine-tuning large language models (LLMs) by approximating updates with low-rank matrices. However, low-rank approximation in two-dimensional space fails to capture high-dimensional structures within the target matrix. Recently, tensor decomposition methods have been explored for fine-tuning LLMs, leveraging their ability to extract structured information. Yet, these approaches primarily rely on random initialization, and the impact of initialization on tensor adaptation remains underexplored. In this paper, we reveal that random initialization significantly diverges from the validation loss achieved by full fine-tuning. To address this, we propose Weight-Decomposed Tensor Adaptation (DoTA), which leverages the Matrix Product Operator (MPO) decomposition of pre-trained weights for effective initialization in fine-tuning LLMs. Additionally, we introduce QDoTA, a quantized version of DoTA designed for 4-bit quantization. Experiments on commonsense and arithmetic reasoning tasks show that DoTA outperforms random initialization methods with fewer parameters. QDoTA further reduces memory consumption and achieves comparable performance to DoTA on commonsense reasoning tasks. We will release our code to support future research.

\end{abstract}
\section{Introduction}

\begin{figure}[t]
  \centering
  \begin{minipage}[t]{0.485\linewidth}
    \centering
    \includegraphics[width=\linewidth]{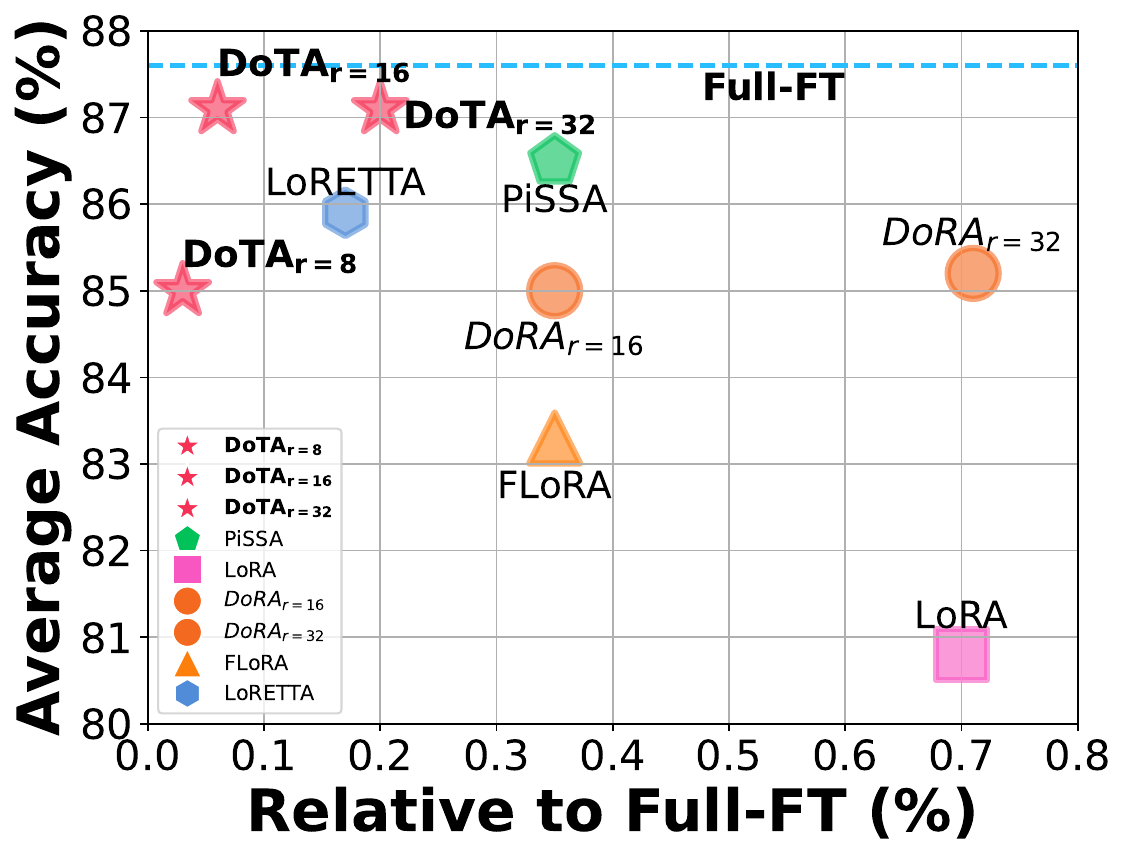}
    \caption{Comparison of the number of trainable parameters and performance across different methods on commonsense reasoning tasks using the LLaMA3-8B model.}
    \label{fig:commonsense-llama3-comparison}
  \end{minipage}%
  \hfill
  \begin{minipage}[t]{0.475\linewidth}
    \centering
    \includegraphics[width=\linewidth]{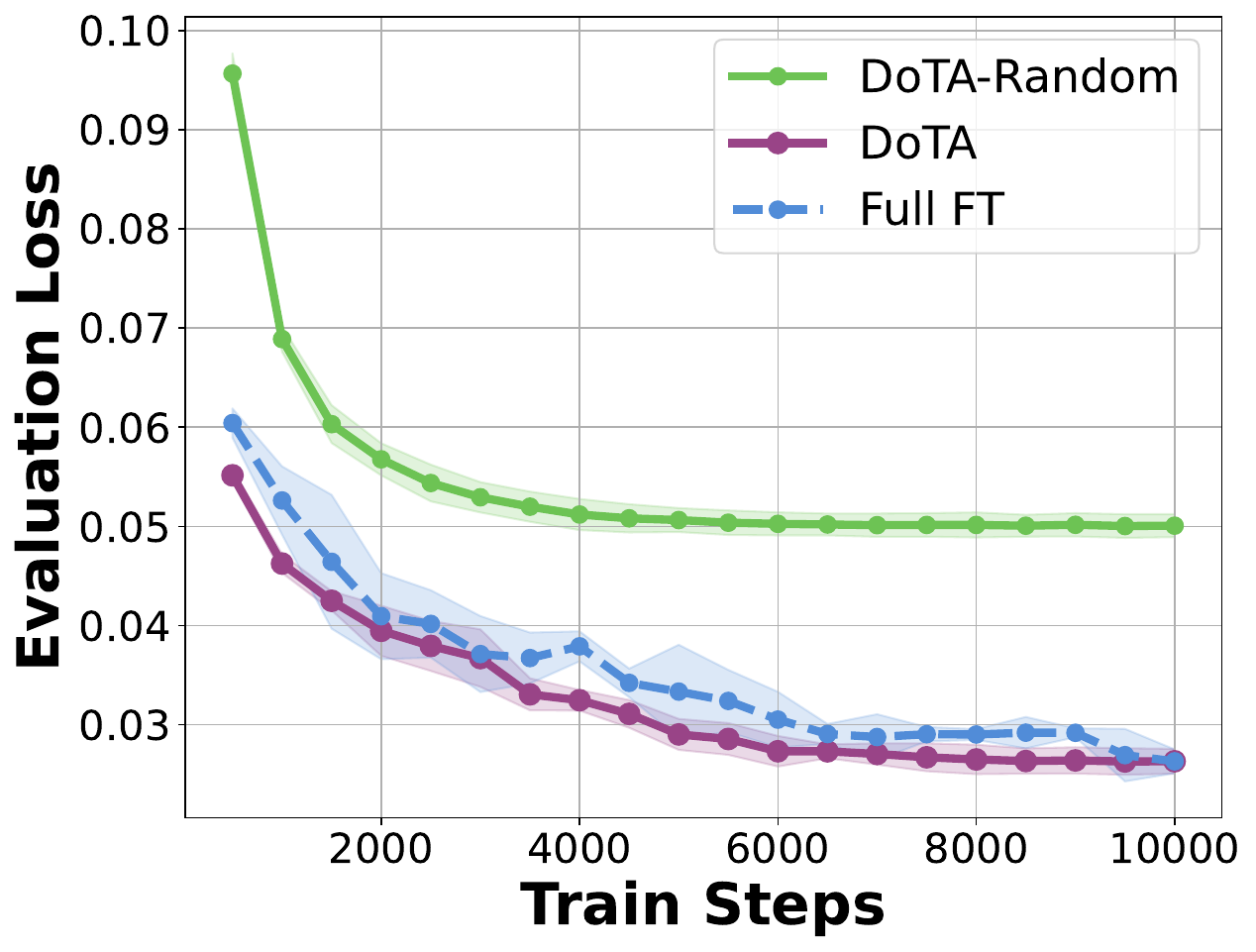}
    \caption{Impact of different initialization methods on evaluation loss. ``DoTA-Random'' indicates we randomly initialized tensors with the same shape as DoTA.}
    \label{fig:random_init}
  \end{minipage}
\end{figure}

Large language models (LLMs) have demonstrated strong performance in NLP tasks \cite{bommasaniOpportunitiesRisksFoundation2022,devlin2018bert,radfordLanguageModelsAre}, but their high computational and memory costs during fine-tuning hinder real-world applications \cite{huangEdgeLLMHighlyEfficient2024}. Parameter-efficient fine-tuning (PEFT) methods, such as Low-Rank Adaptation (LoRA), address these challenges by approximating the updated matrix with two low-rank matrices. However, LoRA and its variants focus on low-rank approximation in two-dimensional space \cite{he2021towards,liu2024dora,meng2024pissa}, ignoring the potential high-dimensional structures within the target matrix.

As a generalization of matrix factorization, tensor decomposition methods have emerged as powerful techniques for approximating high-dimensional space and have been widely applied in tasks like image compression \cite{de2008tensor,gaoCompressionImage2024}, neural network compression \cite{qiuComputeBetterSpent2024}, and recommendation systems \cite{yinTTRecTensorTrain2021}. Recently, efforts have been made to apply tensor decomposition to fine-tuning LLMs \cite{yangLoRETTALowRankEconomic2024,siFLoRALowRankCore2024}. However, these methods largely adopt LoRA’s approach of random initialization for tensor adaptation \cite{yangLoRETTALowRankEconomic2024}.

LoRA assumes that fine-tuning operates within a low-dimensional manifold and initializes low-rank adaptations with random noise or zeros \cite{hu2021lora,li2018measuring}. Yet, random initialization creates a new low-dimensional space that neither inherits the knowledge of the base model nor aligns with the pre-trained manifold \cite{meng2024pissa}. Given the potential of tensor decomposition for capturing high-dimensional structures and the limitations of random initialization, a critical question arises: \textbf{how can we design effective initialization strategies to enable high-dimensional structure updates during fine-tuning?}

In this paper, we propose Weight-Decomposed Tensor Adaptation (DoTA) for fine-tuning LLMs. DoTA uses Matrix Product Operator (MPO) from the quantum community to initialize trainable tensor adaptations by decomposing pre-trained weights, effectively capturing high-dimensional structures for fine-tuning. The number and rank of these tensors determine the trainable parameters, enabling adjustable efficiency. Our contributions are summarized as follows:

$\bullet$ We propose DoTA, a PEFT method that leverages MPO decomposition to capture high-dimensional structures in target updates. Compared to random initialization, DoTA achieves validation loss curves closely aligned with full fine-tuning (Figure \ref{fig:random_init}), demonstrating its effectiveness.

$\bullet$ We evaluate DoTA on commonsense and arithmetic reasoning tasks using LLaMA2-7B and LLaMA3-8B models. DoTA, with significantly fewer parameters, outperforms tensor adaptation methods with random initialization (Figure \ref{fig:commonsense-llama3-comparison}) and even surpasses full fine-tuning on some tasks.

$\bullet$ We introduce QDoTA, an extension of DoTA for 4-bit quantization, which significantly reduces memory usage during fine-tuning while maintaining similar performance to DoTA on commonsense reasoning tasks.

\section{Preliminaries}

In this section, we briefly introduce the preliminaries of tensor algebra. 

\subsection{Tensor and Tensor Operations}

\textbf{Order-N Tensor.} A tensor is a generalization of vectors and matrices to higher dimensions \cite{de2008tensor}. An order-N tensor is an N-dimensional array represented as $\bcalT \in \mathbb{R}^{I_1 \times I_2 \times \ldots \times I_N}$, where $I_n$ is the size of the $n$-th mode. Vectors and matrices are special cases of tensors. An order-N tensor requires $N$ indices to access an element, denoted as $\bcalT_{i_1, \ldots, i_N}$.

\textbf{Tensorization and Matricization.} A vector $\bx \in \mathbb{R}^{\left(\prod_{i=1}^N I_i\right)},$ which is a one-dimensional array with $\prod_{i=1}^N I_i$ elements, can be reshaped into a tensor $\mathcal{X} \in \mathbb{R}^{I_1 \times I_2 \times \ldots \times I_N}$ by tensorization. The mode-$n$ matricization of a tensor $\mathcal{X}$ is obtained by reshaping it along the $n$-th mode, denoted as matrix $\mathbf{X} \in \mathbb{R}^{I_n \times (I_1 I_2 \ldots I_{n-1}  I_{n+1} \ldots I_N)}$.

\textbf{Tensor Contraction.} Tensor contraction generalizes matrix multiplication to higher dimensions by summing over index pairs, reducing the mode of the tensor. Denote the contraction operation as $\otimes$, given tensors $\mathcal{X}^{\prime} \in \mathbb{R}^{I_1 \times \ldots \times I_N \times S_1 \times \cdots \times S_K}$ and $\mathcal{X}^{\prime\prime} \in \mathbb{R}^{S_1 \times \cdots \times S_K \times J_1 \times \ldots \times J_M}$, contraction along indices $S_1, \ldots, S_K$ results in a new tensor $\mathcal{X} 
:= \mathcal{X}^{\prime} \otimes \mathcal{X}^{\prime\prime},$ where $ \mathcal{X} \in \mathbb{R}^{I_1 \times \ldots \times I_N \times J_1 \times \ldots \times J_M}$ and $\mathcal{X}_{i_1, \ldots, i_N, j_1, \ldots, j_M} = \sum_{s_1, \ldots, s_K} \bcalT^{\prime}_{i_1, \ldots, i_N, s_1, \ldots, s_K} \bcalT^{\prime\prime}_{s_1, \ldots, s_K, j_1, \ldots, j_M}$. 

\subsection{Matrix Product Operator} 

MPO is introduced in quantum physics, which is used to represent a large matrix by a sequence product of small tensors \cite{pirvu2010matrix,gao2020compressing}. The small tensors are referred to as core tensors and their tensor contraction approximates the original tensor. By utilizing the contraction operation $\otimes$ as denoted above, the MPO decomposition of matrix $ \bW$ can be represented as
\begin{equation*}
 \text{MPO}(\bW) = \bcalT^{(1)} \otimes \bcalT^{(2)} \otimes \cdots \otimes \bcalT^{(N)},
\end{equation*}
where the core tensor $\bcalT^{(k)} \in \mathbb{R}^{R_{k-1} \times I_k \times J_k \times R_k},$ for $k \in [N],$ in which $ \prod_{k=1}^N I_k =I$ and $\prod_{k=1}^N J_k =J.$ The rank $R_k$ of the core tensor is determined by 
\begin{equation}
    \label{eq:rank}
 R_k = \min \textstyle\left(
    \prod_{n=1}^{k} I_n \times J_n, \prod_{n=k+1}^N I_n \times J_n
 \right), 
\end{equation}
and the rank $R_0 = R_N =1.$ 

The standard MPO decomposition process is shown in Algorithm \ref{alg:mpo}\cite{gao2020compressing,pirvu2010matrix}. The core tensors of matrix $\bW$ are computed by iteratively reshaping the matrix and applying Singular Value Decomposition (SVD). At the $k$-th iteration, the matrix $\bW_{k-1}$ from the previous step is reshaped into a matrix of size $[R_{k-1} \times I_k \times J_k, -1]$. SVD is applied to this matrix, resulting in the decomposition: 

$$ \bU \bSigma \bV^T = \text{SVD}(\bW), $$

where $\bU$ contains the left singular vectors, $\bSigma$ is the diagonal matrix of singular values, and $\bV^T$ contains the right singular vectors. This process repeats for subsequent iterations, passing $\bSigma \bV^T$ to the next step for further decomposition. After $N-1$ iterations, the remaining matrix $\bW$ is assigned as the final core tensor. The output is a sequence of core tensors $\{\bcalT^{(k)}\}_{k=1}^N$, which can be contracted to approximate the original matrix.
\section{The Proposed Method}
\begin{wrapfigure}{r}{0.5\linewidth} 
    \vspace{-45pt} 
    \begin{minipage}{1\linewidth} 
        \begin{algorithm}[H] 
            \caption{MPO decomposition of a matrix}
            \label{alg:mpo}
            \begin{algorithmic}[1]
                \Require matrix $\bW_0 \in \mathbb{R}^{I \times J}$, number of core tensors $N$
                \Ensure MPO tensor set $\{\mathcal{T}^{(k)}\}_{k=1}^N$
                \For{$k = 1$ to $N-1$}
                    \State $\bW_{k-1} \overset{\scalebox{0.5}{\text{reshape}}}{\rightarrow}  \bW[R_{k-1} \times I_k \times J_k, -1]$
                    \State $\bU, \bSigma, \bV^T \leftarrow \text{SVD}(\bW)$
                    \State $\bcalT^{(k)}[R_{k-1}, I_k, J_k, R_k] \overset{\scalebox{0.5}{\text{reshape}}}{\leftarrow} \bU[R_{k-1}\times I_k \times J_k,R_k]$
                    \State $\bW_k \leftarrow \bSigma \bV^T$
                \EndFor
                \State $\bcalT^{(N)} \leftarrow \bW$
                \State \Return $\{\bcalT^{(k)}\}_{k=1}^N$
            \end{algorithmic}
        \end{algorithm}
    \end{minipage}
    \vspace{-40pt} 
\end{wrapfigure}
In this Section, we present the proposed method, DoTA, which initializes the tensor adaptation by leveraging the tensor decomposition of the weight matrix of the LLMs. We first introduce the tensor train decomposition, which is a generalization of the matrix decomposition, and then present the tensor adaptation method for finetuning the LLMs. 

\subsection{MPO-based High-Dimensional Structure Approximation}

With the standard MPO decomposition and ranks $\{R_k\}_{k=1}^{N-1}$ determined by Eq.(\ref{eq:rank}), the original matrix $\bW_0$ can be exactly reconstructed by contracting the core tensors $\{\bcalT^{(k)}\}_{k=1}^N$. To capture the high-dimensional structure and reduce the number of trainable parameters, DoTA truncates the ranks of the core tensors to $\{\bar{R}_k\}_{k=1}^{N-1}$ using a threshold $R$. Compared to low-rank adaptation methods, DoTA is more flexible and better at capturing intricate weight matrix structures.

The number of trainable parameters in DoTA depends on the number of core tensors $N$ and the ranks $\{\bar{R}_k\}_{k=1}^{N-1}$. Given the tensor shapes $\{I_k\}_{k=1}^N$, $\{J_k\}_{k=1}^N$, and ranks $\{\bar{R}_k\}_{k=0}^N$, where $\bar{R}_0 = \bar{R}_N =1.$ , the number of trainable parameters $\rho$ is:
\begin{equation*}
    \rho = \textstyle \sum_{k=1}^N \bar{R}_{k-1} I_k J_k \bar{R}_k. 
\end{equation*}
Moreover, when the ranks of the core tensors satisfy $ \bar{R}_k = R,$ $ \rho $ is computed as $ R(I_1 J_1 + I_N J_N) + R^2 \sum_{k=2}^{N-1} I_k J_k.$ Take the pre-trained matrix with shape $ 1024 \times 1024 $ as an example, the number of trainable parameters of DoTA with $ N=5, R=8, I_k = 4 $ and $ J_k =4 $ is approximately $3.3K,$ which is significantly smaller than the number of trainable parameters of the full fine-tuning method. 
Similar to LoRA, DoTA can be applied to all linear layers of LLMs, except for embedding and head layers in transformers.

\subsection{Weight-Decomposed Tensor Adaptation} 

\begin{figure*}[t]
  \centering
  \includegraphics[width=0.7\linewidth]{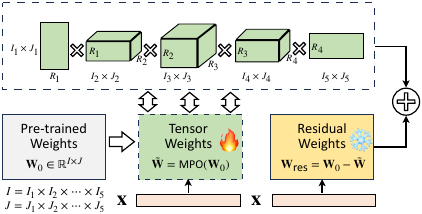} 
  \caption{The architecture of the proposed method. DoTA decomposes the pre-trained weight matrix $\bW_0$ into trainable tensors $\{\bcalT^{(k)}\}_{k=1}^N$ using MPO decomposition. The sequence product of $\{\bcalT^{(k)}\}_{k=1}^N$ reconstruct matrix $\tilde{\bW}$. A residual matrix $ \bW_{\text{res}} $ is formed by subtracting the reconstructed matrix $\tilde{\bW}$ from the original $\bW_0.$} 
  \label{fig:DoTA}
\end{figure*} 

LoRA adapts pre-trained LLMs to new tasks using low-rank matrices \cite{hu2021lora}. Given a pre-trained weight matrix $\bW_0 \in \mathbb{R}^{I \times J}$, the adapted weight matrix $\bW^{\prime}$ is computed as:
\[
\bW^{\prime} = \bW_0 + \bA \bB,
\]
where $\bA \in \mathbb{R}^{I \times R}$ is randomly initialized and $\bB \in \mathbb{R}^{R \times J}$ is initialized to zero. The rank $R$ controls the number of trainable parameters and approximation quality.

DoTA, as illustrated in Figure \ref{fig:DoTA}, adapts pre-trained matrices using a sequence of small tensors $\{\bcalT^{(k)}\}_{k=1}^N$ instead of random initialization. It leverages tensor decomposition of $\bW_0$ for adaptation. Specifically, the tensor adaptation is given by:
\[
\bW^{\prime} = \bW_{\text{res}} + \text{MPO}(\bW_0),
\]
where $\bW_{\text{res}} = \bW_0 - \text{MPO}(\bW_0)$ is the residual matrix, and $\text{MPO}(\bW_0) = \bcalT^{(1)} \otimes \cdots \otimes \bcalT^{(N)}$ is the MPO decomposition of $\bW_0$. During training, the core tensors $\{\bcalT^{(k)}\}_{k=1}^N$ are fine-tuned, while $\bW_{\text{res}}$ is frozen.

DoTA uses a rank threshold $R$ to control the number of trainable parameters by truncating the ranks of the core tensors \cite{gao2020compressing}. To compensate for the information loss due to truncation, it introduces the residual matrix $\bW_{\text{res}}$, which is the difference between the original matrix and the reconstructed matrix from the decomposed tensors. This concept is similar to PiSSA \cite{meng2024pissa}, which also uses a residual matrix to address approximation errors in truncated SVD. 

\subsection{QDoTA: DoTA with Quantization}
QDoTA further reduces memory consumption by quantizing the residual matrices of DoTA using a 4-bit NormalFloat (NF4) data type \cite{dettmers2024qlora} and performing computations with the BFloat16 data type. Given a pre-trained matrix $\bW_0 \in \mathbb{R}^{I \times J}$, the MPO decomposition and residual matrix are computed as in DoTA. The residual matrix is quantized to NF4, and the decomposed core tensors use BFloat16. The forward pass of QDoTA is formulated as:
\[
 \bY = \bX \text{Dequant}\left(\bW_{\text{res}}^{\text{NF4}}\right) + \bX \text{MPO}(\bW_0),
\]
where $\bW_{\text{res}} = \bW_0 - \text{MPO}(\bW_0)$ and $\bW_{\text{res}}^{\text{NF4}}$ is the quantized residual. The activation $\bX$, $\text{MPO}(\bW_0)$, and $\text{Dequant}\left(\bW_{\text{res}}^{\text{NF4}}\right)$ use BFloat16 during training stage.

\section{Experiments}

\begin{table*}[t]
    \caption{This table presents a performance comparison of different models using various methods across 8 commonsense reasoning tasks. The term \#Params (\%) refers to the percentage of trainable parameters. For Llama2-7B, the results for Full Fine-Tuning (Full-FT) are sourced from \cite{chen2024quanta}. The results for LoRA, DoRA, and DoRA$^\dagger$ for Llama2-7B and Llama3-8B are taken from \cite{liu2024dora}. All other results are the mean values across three random seeds. Among all PEFT methods, the best results are highlighted in \textbf{bold}, while the second-best results are \underline{underlined}.}
    \centering
    \resizebox{\textwidth}{!}{
        \begin{tabular}{cccccccccccc}
            \hline
            \textbf{Model} & \textbf{Peft Method} & \textbf{\# Params (\%)} & 
            \textbf{BoolQ} & \textbf{PIQA} & \textbf{SIQA} & \textbf{HellaS.} & \textbf{WinoG.} & \textbf{ARC-e} & \textbf{ARC-c} & \textbf{OBQA} & \textbf{Avg.} \\
            \hline
            \multirow{8}{*}{LLaMA2-7B} 
            & Full-FT & 100\% & 72.9 & 83.0 & 79.8 & 92.4 & 83.0 & 86.6 & 72.0 & 80.1 & 81.2\\
            \cline{2-12}
            & LoRA & 0.83\% & 69.8 & 79.9 & 79.5 & 83.6 & 82.6 & 79.8 & 64.7 & 81.0 & 77.6 \\
            & DoRA & 0.84\% & \underline{71.8} & 83.7 & 76.0 & 89.1 & 82.6 & 83.7 & 68.2 & \underline{82.4} & 79.7 \\
            & DoRA$^\dagger$ & 0.43\% & \textbf{72.0} & 83.1 & \underline{79.9} & 89.1 & \underline{83.0} & 84.5 & \underline{71.0} & 81.2 & \underline{80.5} \\
            & PiSSA & 0.41\% & 69.1&\underline{83.9}&79.6&\underline{91.4}&81.9&\underline{85.4}&68.7&80.4 & 80.1  \\
            & FLoRA & 0.83\% &65.9&79.6&76.2&81.2&77.3&81.7&64.4&75.5&75.2\\
            & LoRETTA & 0.23\% & 66.6&81.1&77.4&\underline{91.4}&79.0&84.3&67.2&74.3 & 77.7 \\
            & \textbf{DoTA (Ours)} & \textbf{0.15\%} & 70.4&\textbf{84.2}&\textbf{81.3}&\textbf{91.7}&\textbf{84.6}&\textbf{85.9}&\textbf{71.7}&\textbf{82.7} &\textbf{81.6} \\
            \hline
            \multirow{8}{*}{LLaMA3-8B} 
            & Full-FT & 100\% & 76.0 & 90.8 & 81.6 & 96.8 & 89.9 & 93.3 & 83.3 & 89.2 & 87.6 \\
            \cline{2-12}
            & LoRA & 0.70\% & 70.8 & 85.2 & 79.9 & 91.7 & 84.3 & 84.2 & 71.2 & 79.0 & 80.8 \\
            & DoRA & 0.71\% & \textbf{74.6} & \underline{89.3} & 79.9 & 95.5 & 85.6 & 90.5 & 80.4 & 85.8 & 85.2 \\
            & DoRA$^\dagger$ & 0.35\% & \underline{74.5} & 88.8 & 80.3 & 95.5 & 84.7 & 90.1 & 79.1 & 87.2 & 85.0 \\
            & PiSSA & 0.35\% &73.8&89.1&\underline{81.9}&\underline{95.8}&\underline{88.4}&\underline{92.6}&\underline{82.4}&\underline{87.9}&\underline{86.5} \\
            & FLoRA & 0.35\% &72.5&86.4&78.9&93.5&83.7&90.0&78.0&83.2 & 83.3\\
            & LoRETTA & 0.17\% &72.7&89.1&81.7&95.7&87.6&\underline{92.6}&82.1&85.5& 85.9 \\  
            & \textbf{DoTA (Ours)} & \textbf{0.06\%} & 74.0&\textbf{89.8}&\textbf{82.8}&\textbf{96.3}&\textbf{88.9}&\textbf{92.9}&\textbf{83.2}&\textbf{89.0} &\textbf{87.1} \\
            \hline
        \end{tabular}
    }

    \label{tab:performance_commonsense}
\end{table*}

We conducted experiments on commonsense and mathematical reasoning tasks using two models: LLaMA2-7B \cite{touvron2023llama2}, and LLaMA3-8B \cite{dubey2024llama3}. (see Appendix \ref{datasets introduction} for a detailed description of the datasets).
Specifically, we first compared DoTA with other parameter-efficient fine-tuning methods including  PiSSA \cite{meng2024pissa}, LoRA \cite{hu2021lora}, DoRA \cite{liu2024dora}, FLoRA\cite{siFLoRALowRankCore2024} and LoRETTA\cite{yangLoRETTALowRankEconomic2024} as well as full-parameter fine-tuning on commonsense and mathematical reasoning tasks. We then evaluated these methods under quantization settings to demonstrate DoTA's effectiveness in reducing quantization errors. Additionally, ablation studies were conducted to validate the necessity of DoTA's initialization approach compared to random initialization. Finally, we examined the impact of different ranks of the core tensor on DoTA, highlighting the trade-off between parameter count and performance. Unless otherwise specified, all the results we report are the mean values obtained from three different random seeds.

For our method, DoTA, we conducted an extensive hyperparameter search over \( N \in \{3, 5, 7, 9\} \) and \( R \in \{16, 32, 64, 128\} \), ultimately selecting \( N = 5 \) and \( R = 16 \) unless specified otherwise, as they consistently delivered the best performance. Tensor shapes $ \{I_k\}_{k=1}^N, \{J_k\}_{k=1}^N$ and the remaining hyperparameter settings used in the experiments are detailed in Appendix \ref{experiment settings}.

\subsection{Commonsense Reasoning}

The model was fine-tuned on the Commonsense-170K dataset \cite{hu2023llm-adapters} and subsequently evaluated on the individual test sets for each sub-task.

\textbf{DoTA outperforms other baseline methods across two models}. As shown in Table \ref{tab:performance_commonsense}, our method outperforms other baseline methods while using fewer trainable parameters, except for the full fine-tuning method on LLaMA3-8B. On LLaMA2-7B, our method outperforms the alternatives by a margin of 1.1\% to 6.4\%. Furthermore, on LLaMA3-8B, our method leads other parameter-efficient fine-tuning methods by an average accuracy margin of 0.6\% to 6.3\%, while using fewer parameters.

These results demonstrate that our method effectively captures the most crucial aspects of the original weight matrix, achieving superior results with minimal modifications. Although our method falls just 0.5\% short of full fine-tuning on LLaMA3-8B, it utilizes only 0.06\% of the parameters, underscoring its significant practical advantages.

\subsection{Mathematical Reasoning}

\begin{wrapfigure}{r}{0.455\linewidth} 
    \vspace{-20pt}
    \centering
    \includegraphics[width=\linewidth]{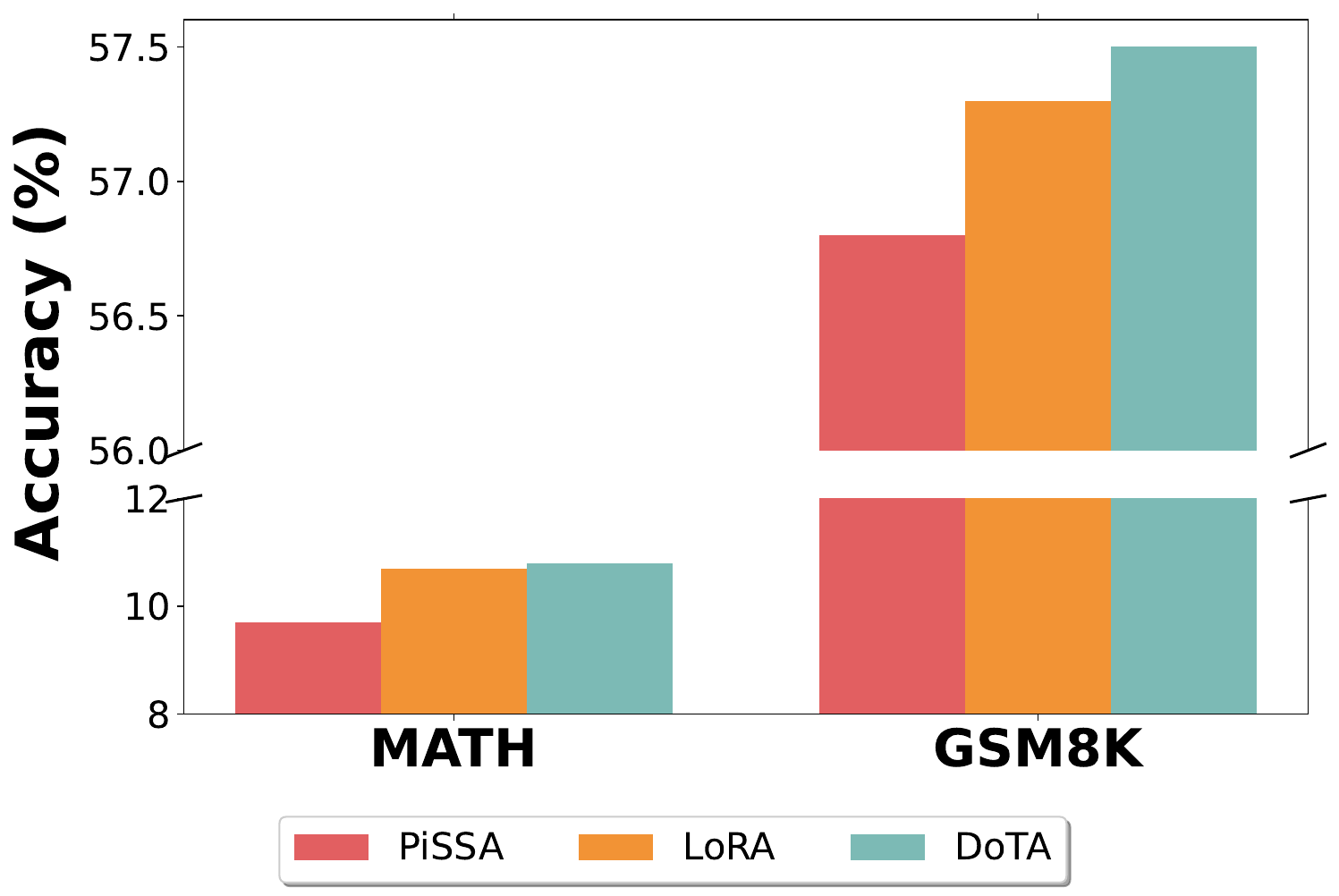}
    \caption{Performance of different methods on mathematical reasoning tasks using the LLaMA2-7B model}
    \label{fig:mathematical-llama2}
    \vspace{-25pt}
\end{wrapfigure}

We fine-tuned LLaMA2-7B on the MetaMathQA-395K \cite{yu2023metamath} dataset to enhance its mathematical capabilities and evaluated its performance on the MATH \cite{yu2023metamath} and GSM8K \cite{gsm8k} datasets. We also tested FLoRA and LoRETTA on these two datasets, but since their performance was inferior to PiSSA, their results are not shown in the figures. All results are reported as the average across three random seeds.

\textbf{DoTA shows more significant improvement on GSM8K}. As shown in Figure \ref{fig:mathematical-llama2}, our method performs slightly better than the others on the MATH dataset, though the improvement is modest. We attribute this to the model's limited capabilities acquired during the pre-training phase, which may not be sufficient for this task. However, on the GSM8K dataset, our method achieves a significant improvement over others, further demonstrating its ability to leverage structural information within the weight matrix for more effective fine-tuning.

\subsection{Quantitative Analysis}

In this section, we compare DoTA with PiSSA and LoRA under quantization settings, referred to as QDoTA, QPiSSA \cite{meng2024pissa}, and QLoRA \cite{dettmers2024qlora}, using the LLaMA3-8B model for commonsense reasoning tasks.

\begin{figure*}[t]
    \centering
    \includegraphics[width=0.90\linewidth]{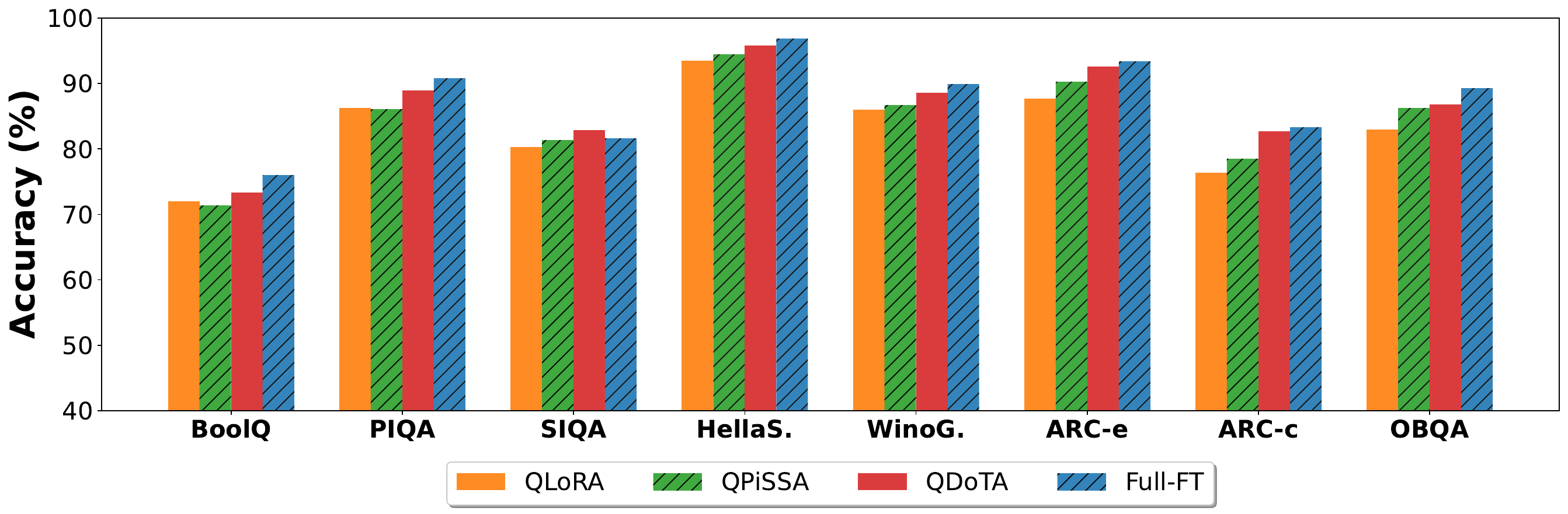}
    \caption{Comparison of the quantized versions of various methods on eight commonsense reasoning tasks using the LLaMA3-8B model. QPiSSA and QLoRA use 0.7\% of the parameters required for full fine-tuning, while DoTA uses 0.2\%.}
    \label{fig:Commonsense-llama3-Quantization}
\end{figure*}

\textbf{QDoTA reduces quantization error to improve performance}. Figure \ref{fig:Commonsense-llama3-Quantization} shows that our method consistently outperforms QLoRA and QPiSSA across all subtasks of commonsense reasoning. It is important to note that full fine-tuning was not subjected to any quantization, meaning it did not experience quantization errors. Nevertheless, our method, even under quantized settings, remains close to or on par with full fine-tuning. This demonstrates that our method is more effective than other PEFT approaches in mitigating quantization errors.

\subsection{Ablation Study}

In this section, we compare tensors randomly initialized with those decomposed from the original matrix to demonstrate our method's efficiency. For random initialization, we apply Gaussian initialization to all tensors from each matrix decomposition, except for the last tensor, which is initialized to zero. We collect results from three random seeds and plot the mean and standard deviation of the evaluation loss for Llama3-8B on the Commensence Reasoning task, comparing different initialization methods.

\textbf{DoTA's initialization method is crucial}. Figure \ref{fig:random_init} shows that the randomly initialized tensor tends to converge to a suboptimal point during training, where the validation loss stagnates. In contrast, the curve for DoTA closely mirrors that of full fine-tuning, exhibiting a downward trend with fluctuations and ultimately converging to a more optimal point, very close to the performance of full fine-tuning. We attribute this to DoTA's use of decomposed tensors, which preserve more of the original matrix's information, allowing for more effective fine-tuning.

\subsection{Analyzing the Impact of Rank}
Lastly, we analyzed the impact of varying hyperparameter ranks on DoTA's performance. We investigated the impact of different ranks [8, 16, 32] on performance using LLaMA2-7B and LLaMA3-8B across the previously defined tasks. 
\begin{figure*}[htbp]
    \centering
    \begin{subfigure}{0.325\linewidth}
        \includegraphics[width=\linewidth]{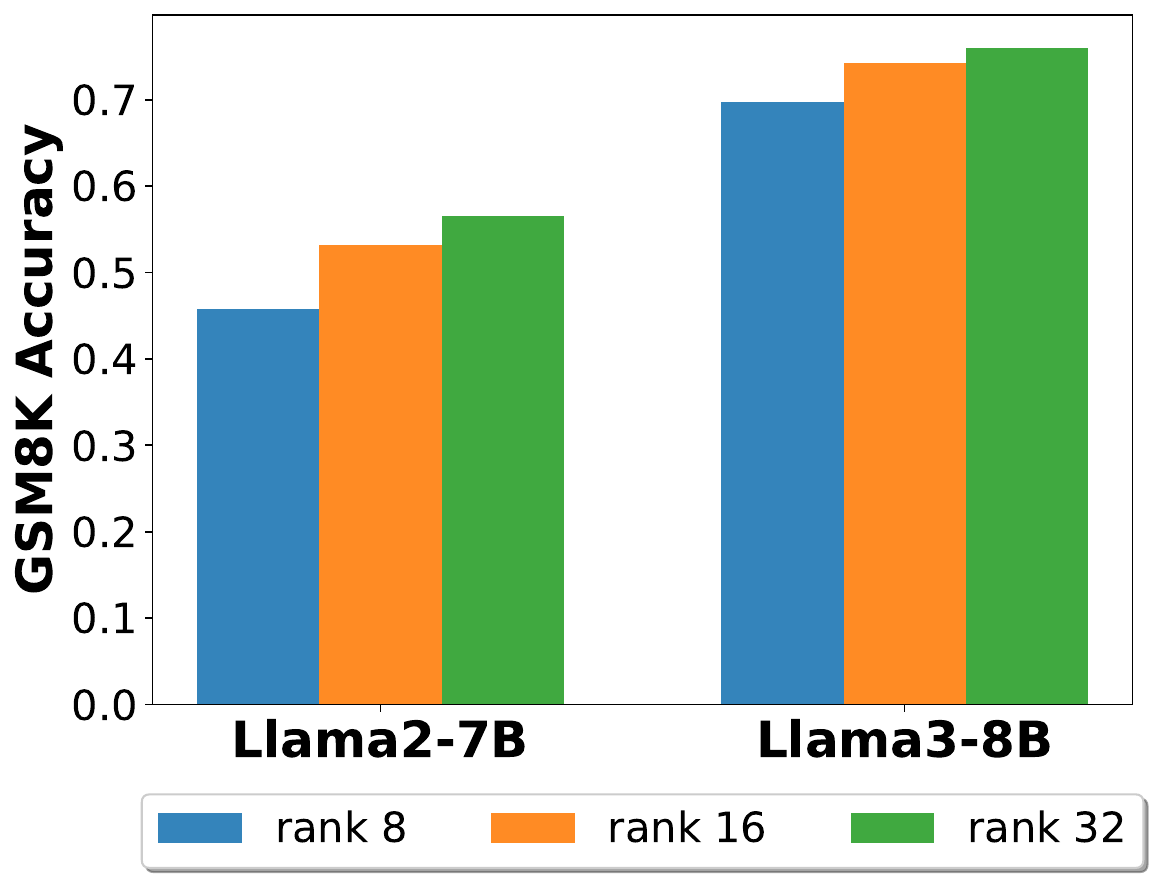}
        \caption{GSM8K}
        \label{fig:GSM8K_diff_rank}
    \end{subfigure}
    \hfill
    \begin{subfigure}{0.325\linewidth}
        \includegraphics[width=\linewidth]{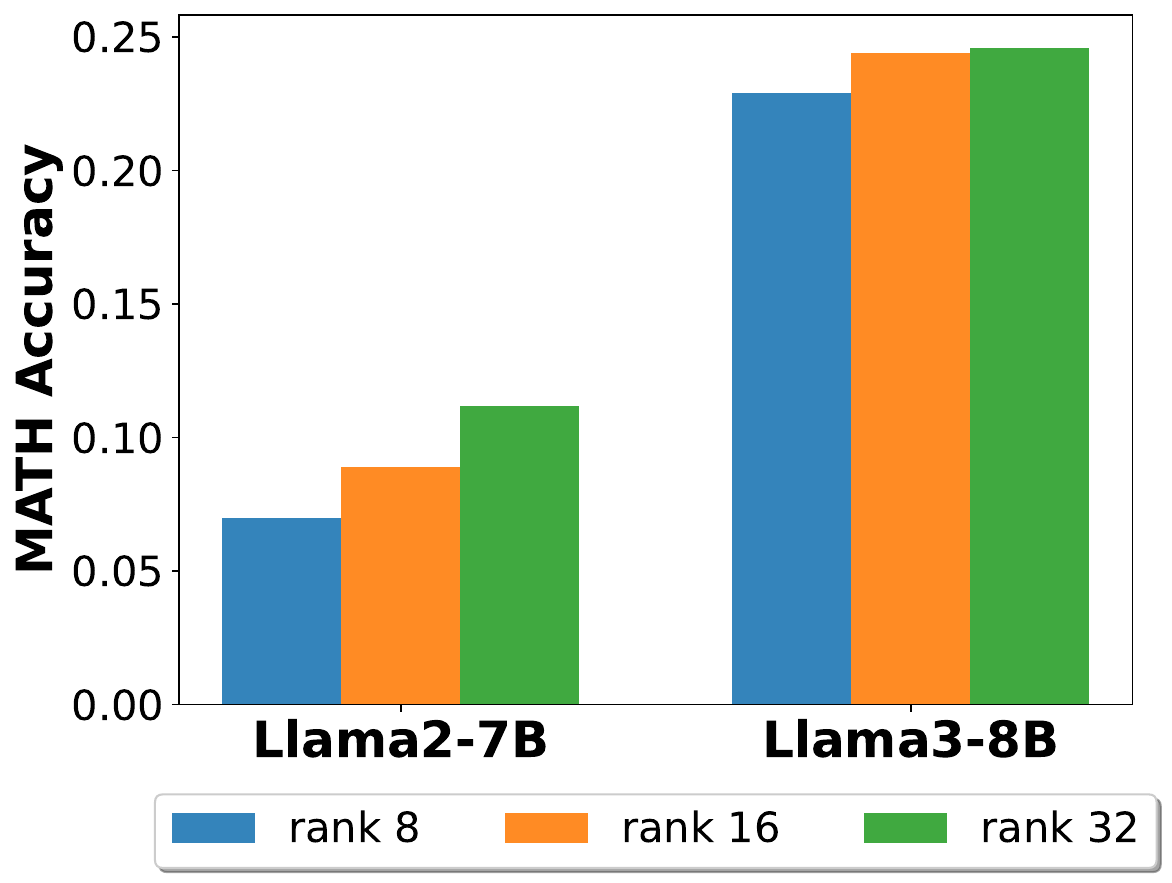}
        \caption{MATH}
        \label{fig:MATH_diff_rank}
    \end{subfigure}
    \hfill
    \begin{subfigure}{0.325\linewidth}
        \includegraphics[width=\linewidth]{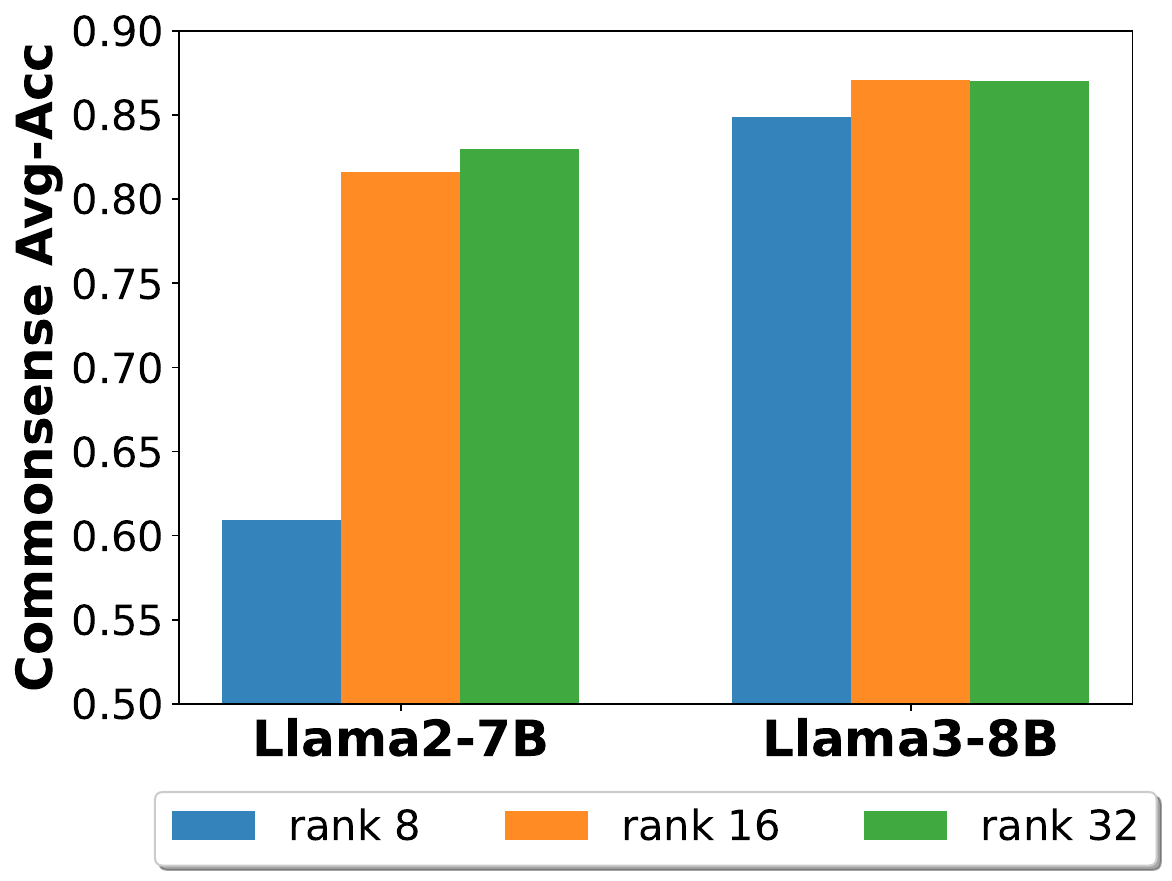}
        \caption{Commonsense}
        \label{fig:Commonsense_diff_rank}
    \end{subfigure}
    \caption{Accuracy across different ranks for various tasks.}
    \label{fig:diff_rank}
\end{figure*}

\textbf{Lower rank may reduce performance}. Lower ranks generally lead to worse test results, especially in mathematical reasoning tasks (Figures \ref{fig:GSM8K_diff_rank} and \ref{fig:MATH_diff_rank}), and rank 8 underperforms in commonsense reasoning tasks on LLaMA2-7B (Figure \ref{fig:Commonsense_diff_rank}). This is due to the method's sensitivity to task and model structure, where a lower rank may fail to capture sufficient information, causing instability.

\textbf{Higher rank does not always improve performance}. The difference between rank 16 and rank 32 is minimal, particularly for commonsense reasoning on LLaMA3-8B (Figure \ref{fig:Commonsense_diff_rank}), where performance is nearly identical. These findings suggest that while low ranks can hinder training, higher ranks do not always yield substantial gains. Thus, rank selection for DoTA requires balancing performance and memory usage.

\section{Related Work}

\textbf{Parameter-Efficient Fine-Tuning} To adapt LLMs to downstream tasks with limited resources, PEFT methods like LoRA have been proposed to reduce learnable parameters \cite{devlin2018bert,huangEdgeLLMHighlyEfficient2024}. LoRA approximates weight changes with low-rank matrices \cite{hu2021lora}, and follow-up works enhance efficiency and effectiveness \cite{kalajdzievski2023rank}. LoRA+ \cite{hayouLoRAEfficientLow2024} uses different learning rates for low-rank matrices to improve convergence. AdaLoRA \cite{zhangAdaLoRAAdaptiveBudget2023} adapts the rank of these matrices during fine-tuning, while rsLoRA \cite{kalajdzievski2023rank} modifies scaling factors for stability. Additionally, structure matrices are introduced to boost computational efficiency and model representation \cite{qiuComputeBetterSpent2024}.

\textbf{Low-Rank Adaptation Initilization} Recent literature shows that initializing low-rank matrices with pre-trained weight decomposition can improve fine-tuning efficiency \cite{liu2024dora,meng2024pissa,wangLoRAGALowRankAdaptation2024}. DoRA \cite{liu2024dora} decomposes pre-trained weights into magnitude and direction, enhancing learning capacity and stability. PiSSA \cite{meng2024pissa} uses singular vectors of the pre-trained weight for faster convergence and better performance. LoRA-GA \cite{wangLoRAGALowRankAdaptation2024} aligns the gradients of low-rank matrix products with those of full fine-tuning at the first step. MiLoRA \cite{wangMiLoRAHarnessingMinor2024} updates minor singular components while preserving principal ones, maintaining pre-trained knowledge for superior performance.

\textbf{Tensor Adaptation for Fine-tuning} Tensor decomposition methods represent high-dimensional data as combinations of small tensors, useful for compressing datasets and models \cite{de2008tensor,gao2020compressing,yinTTRecTensorTrain2021}. Several works apply tensor decomposition for fine-tuning language models \cite{siFLoRALowRankCore2024,yangLoRETTALowRankEconomic2024,chen2024quanta,bershatsky2024lotr}. LoRETTA \cite{yangLoRETTALowRankEconomic2024} uses tensor-train decomposition to improve multi-task learning and anti-overfitting. FLoRA \cite{siFLoRALowRankCore2024} applies Tucker decomposition for low-rank tensor adaptations, enhancing fine-tuning efficiency. QuanTA \cite{chen2024quanta} introduces a quantum-inspired method for efficient high-rank representations, outperforming low-rank approaches. 
\section{Conclusion}
This paper introduces DoTA, a weight-decomposed tensor adaptation method for fine-tuning LLMs, which retains pre-trained knowledge through tensor decomposition. DoTA is simple, effective, and easily integrable, outperforming existing methods with fewer trainable parameters in commonsense and mathematical reasoning tasks. We also present QDoTA, a 4-bit quantized version that reduces memory costs while outperforming other 4-bit quantized methods in commonsense reasoning. Ablation studies emphasize the importance of DoTA's initialization method for its performance. DoTA offers an efficient solution for adapting LLMs to new tasks with limited computational resources. 

\appendix
\section{Datasets introduction}
\label{datasets introduction}

Commonsense reasoning tasks encompass eight subtasks: BoolQ, PIQA, SIQA, HellaSwag, Winogrande, ARC-e and ARC-c, and OBQA. Following the methodology of \cite{hu2023llm-adapters} and \cite{liu2024dora}, they constructed the Commonsense-170K which is a fine-tuning dataset by combining the training sets from these subtasks. 

The MetaMathQA-395K dataset\cite{yu2023metamath} consists of 395,000 mathematically diverse questions. The GSM8K dataset \cite{gsm8k} comprises elementary-level math problems, while the MATH dataset \cite{yu2023metamath} includes more complex high school competition and exam questions.
\section{Experiment details}

\label{experiment settings}

Following DoRA \cite{liu2024dora} and PiSSA \cite{meng2024pissa}, we limited the commonsense reasoning dataset to 100,000 samples, while using all available samples for mathematical reasoning. Each dataset was split into a training set and a 1,200-sample validation set. 
All experiments used a training epoch of 1, with method-specific learning rates to optimize performance. The model achieving the lowest validation loss was saved for downstream evaluation.
To ensure consistency, all models were initialized with their original pre-trained precision and utilized bf16 for computation.

\begin{wraptable}{r}{0.5\textwidth} 
    \vspace{-28pt}
    \centering
    \begin{tabular}{cc}
        \toprule
        \textbf{Hidden dimension} & \textbf{Tensor shape} \\
        \midrule
        50400 & [5, 10, 14, 12, 6] \\
        14336 & [4, 8, 8, 8, 7] \\
        11008 & [4, 4, 43, 4, 4] \\
        768   & [4, 4, 4, 4, 3] \\
        3072  & [4, 4, 8, 6, 4] \\
        1024  & [4, 4, 4, 4, 4] \\
        4096  & [4, 4, 8, 8, 4] \\
        2304  & [4, 4, 8, 6, 3] \\
        \bottomrule
    \end{tabular}
    \caption{Tensor shape for hidden dimension}
    \label{default shape}
    \vspace{-50pt}
\end{wraptable}

For LoRA and LoRETTA, we set \(r = 32\) (with \(\alpha = r\)) and use a learning rate of \(1 \times 10^{-4}\). For PiSSA, \(r = 16\) (with \(\alpha = r\)) and the same learning rate of \(1 \times 10^{-4}\) are applied to both models. FLoRA employs \(r = 32\) for LLaMA2-7B and \(r = 16\) for LLaMA3-8B. For FLoRA and LoRETTA, we adopt the other default parameters specified in their respective papers \cite{yangLoRETTALowRankEconomic2024,siFLoRALowRankCore2024}. Full-FT adjusts all parameters with a learning rate of 1e-5.

All experiments were conducted under consistent settings: AdamW was used as the optimizer, with a cosine learning rate scheduler, a batch size of 16, a warmup ratio of 0.03, no dropout, and bf16 precision. The components \(Q\), \(K\), \(V\), \(U\), and \(D\) correspond to \texttt{q\_proj}, \texttt{k\_proj}, \texttt{v\_proj}, \texttt{up\_proj}, and \texttt{down\_proj}, respectively. These layers served as adapters for LLaMA2-7B and LLaMA3-8B across all methods unless explicitly stated otherwise. Tensor shapes \(\{I_k\}_{k=1}^N\) and \(\{J_k\}_{k=1}^N\) are detailed in Table~\ref{default shape}. All experiments were performed on an NVIDIA A100-80GB GPU.

\textbf{Experiment settings for Commonsense Reasoning}
For LoRA and LoRETTA, we set \(r = 32\) (with \(\alpha = r\)) and use a learning rate of $1 \times 10^{-4}.$ For PiSSA, \(r = 16\) (with \(\alpha = r\)) and the same learning rate of $1 \times 10^{-4}$ are applied to both models. FLoRA employs \(r = 32\) for LLaMA2-7B and \(r = 16\) for LLaMA3-8B. For FLoRA and LoRETTA, we adopt the other default parameters specified in their respective papers \cite{yangLoRETTALowRankEconomic2024,siFLoRALowRankCore2024}. Full-FT adjusts all parameters with a learning rate of 1e-5.

\textbf{Experiment settings for Mathematical Reasoning} LoRA and PiSSA share a rank \(r = 32\) (with \(\alpha = r\)), while DoTA uses a tensor-number \( N \) of 5 and tensor-rank \( R \) of 32. Learning rates are \( 3\times 10^{-4} \) for LoRA and DoTA, \( 1\times 10^{-4} \) for PiSSA, and \( 1\times 10^{-5} \) for full fine-tuning.  

\textbf{Experiment settings for Quantization} All methods use NF4 quantification, with a tensor-rank \( R = 32 \). For QDoTA, the learning rate is \( 1\times 10^{-4} \) for commonsense tasks. For QLoRA, the learning rate is \( 1\times 10^{-5} \). For QPiSSA, the learning rate is \( 1\times 10^{-4} \). LLaMA3-8B was used for commonsense tasks.

\bibliographystyle{splncs04}
\bibliography{samplepaper} 

\begin{thebibliography}{10}
\providecommand{\url}[1]{\texttt{#1}}
\providecommand{\urlprefix}{URL }
\providecommand{\doi}[1]{https://doi.org/#1}

\bibitem{bershatsky2024lotr}
Bershatsky, D., Cherniuk, D., Daulbaev, T., Oseledets, I.: Lotr: Low tensor rank weight adaptation. arXiv preprint arXiv:2402.01376  (2024)

\bibitem{bommasaniOpportunitiesRisksFoundation2022}
Bommasani, R., Hudson, D.A., Adeli, E., Altman, e.a.: On the {Opportunities} and {Risks} of {Foundation} {Models} (Jul 2022), arXiv:2108.07258 [cs]

\bibitem{chen2024quanta}
Chen, Z., Dangovski, R., Loh, C., Dugan, O., Luo, D., Solja{\v{c}}i{\'c}, M.: Quanta: Efficient high-rank fine-tuning of llms with quantum-informed tensor adaptation. arXiv preprint arXiv:2406.00132  (2024)

\bibitem{gsm8k}
Cobbe, K., Kosaraju, V., Bavarian, M., Chen, M., Jun, H., Kaiser, L., Plappert, M., Tworek, J., Hilton, J., Nakano, R., et~al.: Training verifiers to solve math word problems. arXiv preprint arXiv:2110.14168  (2021)

\bibitem{de2008tensor}
De~Silva, V., Lim, L.H.: Tensor rank and the ill-posedness of the best low-rank approximation problem. SIAM Journal on Matrix Analysis and Applications  \textbf{30}(3),  1084--1127 (2008)

\bibitem{dettmers2024qlora}
Dettmers, T., Pagnoni, A., Holtzman, A., Zettlemoyer, L.: Qlora: Efficient finetuning of quantized llms. Advances in Neural Information Processing Systems  \textbf{36} (2024)

\bibitem{devlin2018bert}
Devlin, J., Chang, M., Lee, K., Toutanova, K.: {BERT:} pre-training of deep bidirectional transformers for language understanding. In: Burstein, J., Doran, C., Solorio, T. (eds.) Proceedings of the 2019 Conference of the North American Chapter of the Association for Computational Linguistics: Human Language Technologies, {NAACL-HLT} 2019, Minneapolis, MN, USA, June 2-7, 2019, Volume 1 (Long and Short Papers). pp. 4171--4186. Association for Computational Linguistics (2019)

\bibitem{dubey2024llama3}
Dubey, A., Jauhri, A., Pandey, A., Kadian, A., Al-Dahle, A., Letman, A., Mathur, A., Schelten, A., Yang, A., Fan, A., et~al.: The llama 3 herd of models. arXiv preprint arXiv:2407.21783  (2024)

\bibitem{gao2020compressing}
Gao, Z.F., Cheng, S., He, R.Q., Xie, Z., Zhao, H.H., Lu, Z.Y., Xiang, T.: Compressing deep neural networks by matrix product operators. Physical Review Research  \textbf{2}(2),  023300 (2020)

\bibitem{gaoCompressionImage2024}
Gao, Z.F., Liu, P., Zhao, W.X., Xie, Z.Y., Wen, J.R., Lu, Z.Y.: Compression image dataset based on multiple matrix product states. In: Arai, K. (ed.) Advances in information and communication. pp. 621--638. Springer Nature Switzerland, Cham (2024)

\bibitem{hayouLoRAEfficientLow2024}
Hayou, S., Ghosh, N., Yu, B.: {LoRA}+: {Efficient} {Low} {Rank} {Adaptation} of {Large} {Models} (Jul 2024), arXiv:2402.12354 [cs, stat]

\bibitem{he2021towards}
He, J., Zhou, C., Ma, X., Berg-Kirkpatrick, T., Neubig, G.: Towards a unified view of parameter-efficient transfer learning. arXiv preprint arXiv:2110.04366  (2021)

\bibitem{hu2021lora}
Hu, E.J., Shen, Y., Wallis, P., Allen-Zhu, Z., Li, Y., Wang, S., Wang, L., Chen, W.: Lora: Low-rank adaptation of large language models. arXiv preprint arXiv:2106.09685  (2021)

\bibitem{hu2023llm-adapters}
Hu, Z., Wang, L., Lan, Y., Xu, W., Lim, E.P., Bing, L., Xu, X., Poria, S., Lee, R.K.W.: Llm-adapters: An adapter family for parameter-efficient fine-tuning of large language models. arXiv preprint arXiv:2304.01933  (2023)

\bibitem{huangEdgeLLMHighlyEfficient2024}
Huang, M., Shen, A., Li, K., Peng, H., Li, B., Yu, H.: {EdgeLLM}: {A} {Highly} {Efficient} {CPU}-{FPGA} {Heterogeneous} {Edge} {Accelerator} for {Large} {Language} {Models} (Jul 2024), arXiv:2407.21325 [cs]

\bibitem{kalajdzievski2023rank}
Kalajdzievski, D.: A rank stabilization scaling factor for fine-tuning with lora. arXiv preprint arXiv:2312.03732  (2023)

\bibitem{li2018measuring}
Li, C., Farkhoor, H., Liu, R., Yosinski, J.: Measuring the intrinsic dimension of objective landscapes. arXiv preprint arXiv:1804.08838  (2018)

\bibitem{liu2024dora}
Liu, S.Y., Wang, C.Y., Yin, H., Molchanov, P., Wang, Y.C.F., Cheng, K.T., Chen, M.H.: Dora: Weight-decomposed low-rank adaptation. arXiv preprint arXiv:2402.09353  (2024)

\bibitem{meng2024pissa}
Meng, F., Wang, Z., Zhang, M.: Pissa: Principal singular values and singular vectors adaptation of large language models. arXiv preprint arXiv:2404.02948  (2024)

\bibitem{pirvu2010matrix}
Pirvu, B., Murg, V., Cirac, J.I., Verstraete, F.: Matrix product operator representations. New Journal of Physics  \textbf{12}(2),  025012 (2010)

\bibitem{qiuComputeBetterSpent2024}
Qiu, S., Potapczynski, A., Finzi, M., Goldblum, M., Wilson, A.G.: Compute {Better} {Spent}: {Replacing} {Dense} {Layers} with {Structured} {Matrices} (Jun 2024), arXiv:2406.06248 [cs]

\bibitem{radfordLanguageModelsAre}
Radford, A., Wu, J., Child, R., Luan, D., Amodei, D., Sutskever, I.: Language {Models} are {Unsupervised} {Multitask} {Learners}

\bibitem{siFLoRALowRankCore2024}
Si, C., Wang, X., Yang, X., Xu, Z., Li, Q., Dai, J., Qiao, Y., Yang, X., Shen, W.: {FLoRA}: {Low}-{Rank} {Core} {Space} for {N}-dimension (May 2024), arXiv:2405.14739 [cs]

\bibitem{touvron2023llama2}
Touvron, H., Martin, L., Stone, K., Albert, P., Almahairi, A., Babaei, Y., Bashlykov, N., Batra, S., Bhargava, P., Bhosale, S., et~al.: Llama 2: Open foundation and fine-tuned chat models. arXiv preprint arXiv:2307.09288  (2023)

\bibitem{wangMiLoRAHarnessingMinor2024}
Wang, H., Xiao, Z., Li, Y., Wang, S., Chen, G., Chen, Y.: {MiLoRA}: {Harnessing} {Minor} {Singular} {Components} for {Parameter}-{Efficient} {LLM} {Finetuning} (Jun 2024), arXiv:2406.09044 [cs]

\bibitem{wangLoRAGALowRankAdaptation2024}
Wang, S., Yu, L., Li, J.: {LoRA}-{GA}: {Low}-{Rank} {Adaptation} with {Gradient} {Approximation} (Jul 2024), arXiv:2407.05000 [cs]

\bibitem{yangLoRETTALowRankEconomic2024}
Yang, Y., Zhou, J., Wong, N., Zhang, Z.: {LoRETTA}: {Low}-{Rank} {Economic} {Tensor}-{Train} {Adaptation} for {Ultra}-{Low}-{Parameter} {Fine}-{Tuning} of {Large} {Language} {Models} (Feb 2024), arXiv:2402.11417 [cs]

\bibitem{yinTTRecTensorTrain2021}
Yin, C., Acun, B., Liu, X., Wu, C.J.: {TT}-{Rec}: {Tensor} {Train} {Compression} for {Deep} {Learning} {Recommendation} {Models} (Jan 2021), arXiv:2101.11714 [cs]

\bibitem{yu2023metamath}
Yu, L., Jiang, W., Shi, H., Yu, J., Liu, Z., Zhang, Y., Kwok, J.T., Li, Z., Weller, A., Liu, W.: Metamath: Bootstrap your own mathematical questions for large language models. arXiv preprint arXiv:2309.12284  (2023)

\bibitem{zhangAdaLoRAAdaptiveBudget2023}
Zhang, Q., Chen, M., Bukharin, A., Karampatziakis, N., He, P., Cheng, Y., Chen, W., Zhao, T.: {AdaLoRA}: {Adaptive} {Budget} {Allocation} for {Parameter}-{Efficient} {Fine}-{Tuning} (Mar 2023)

\end{thebibliography}
\end{document}